\pdfoutput=1

\documentclass[11pt]{article}

\usepackage[final]{acl}

\usepackage{blindtext}
\usepackage{hyperref}
\usepackage{enumitem}

\usepackage{times}
\usepackage{latexsym}

\usepackage[T1]{fontenc}

\usepackage[utf8]{inputenc}

\usepackage{microtype}

\usepackage{inconsolata}

\usepackage{graphicx}
\usepackage{booktabs}
\usepackage{mathtools} 
\usepackage{amsmath}
\usepackage{xcolor}

\title{Use Random Selection for Now: Investigation of Few-Shot Selection Strategies in LLM-based Text Augmentation for Classification}

\author{Jan Cegin$^{\spadesuit}$$^\dagger$, Branislav Pecher$^{\spadesuit}$$^\dagger$, Jakub Simko$^\dagger$, Ivan Srba$^\dagger$ Maria Bielikova$^\dagger$,
{\bf Peter Brusilovsky}$^\ddagger$ \\
  $^{\spadesuit}$ Faculty of Information Technology, Brno University of Technology, Brno, Czechia \\
  $^\dagger$ Kempelen Institute of Intelligent Technologies, Bratislava, Slovakia\\
  \texttt{\{jan.cegin, branislav.pecher, jakub.simko, ivan.srba, maria.bielikova\}}@kinit.sk \\
  $^\ddagger$ University of Pittsburgh, Pittsburgh, USA \\
  \texttt{peterb@pitt.edu }}

\begin{document}
\maketitle
\begin{abstract}
The generative large language models (LLMs) are increasingly used for data augmentation tasks, where text samples are paraphrased (or generated anew) and then used for classifier fine-tuning. Existing works on augmentation leverage the few-shot scenarios, where samples are given to LLMs as part of prompts, leading to better augmentations. Yet, the samples are mostly selected randomly and a comprehensive overview of the effects of other (more ``informed'') sample selection strategies is lacking. In this work, we compare sample selection strategies existing in few-shot learning literature and investigate their effects in LLM-based textual augmentation. We evaluate this on in-distribution and out-of-distribution classifier performance. Results indicate, that while some ``informed'' selection strategies increase the performance of models, especially for out-of-distribution data, it happens only seldom and with marginal performance increases. Unless further advances are made, a default of random sample selection remains a good option for augmentation practitioners.
\end{abstract}

\section{Introduction}

The emergence of recent large language models (LLMs) such as GPT-4, Gemini, Llama, and their wide availability, prompted their use in \emph{augmentation} of textual datasets~\cite{ubani2023zeroshotdataaug, dai2023auggpt, piedboeuf-langlais-2023-chatgpt, cegin-etal-2023-chatgpt, cegin2024effectsdiversityincentivessample}. LLM augmentation has been used in various domains such as sentiment analysis~\cite{ONAN2023101611, piedboeuf-langlais-2023-chatgpt}, intent classification~\cite{cegin-etal-2023-chatgpt}, news classification~\cite{piedboeuf-langlais-2023-chatgpt, cegin2024effectsdiversityincentivessample}, and health symptoms classification~\cite{dai2023auggpt}. In most LLM-based augmentation scenarios, the dataset size is increased through \emph{paraphrasing} of original samples or \emph{generation} of completely new samples that adhere a specified label. This can be done without any samples provided (\textit{zero-shot}). Alternatively, one can include already existing samples as part of the prompt to better instruct the LLM {few-shot}). The augmented datasets are then used for training \emph{downstream} classifiers, which are usually much smaller than the prompted LLMs, and thus cheaper and more suitable for production environments.

Recent studies report better performance for few-shot LLM-based augmentation, as compared with zero-shot approaches for text classification~\cite{cegin2024effectsdiversityincentivessample, piedboeuf2024evaluationprotocolsdataaugmentation}. At the same time, most existing few-shot augmentation studies select the samples randomly, and the potential of using more \textit{informed} selection strategies (existing elsewhere in few-shot learning literature) is under-explored. Furthermore, augmentation studies mostly focus on paraphrasing and and are evaluated in-distribution.


In few-shot learning, the \textit{informed} sample selection strategies aim to select most relevant samples that would lead to better outputs. The samples can be selected based on their similarity, diversity, informativeness, or quality~\cite{li-qiu-2023-finding, zhang-etal-2022-active, chang-jia-2023-data, pecher2024automatic}. Through these methods, LLMs can potentially produce better augmentations, in return for the additional computation costs of the informed sample selection. \textit{Literature shows, that the choice of samples for few-shot learning significantly influences its outcomes} (i.e. sensitivity of sample selection)~\cite{pecher-survey-2024, zhang-etal-2022-active, koksal-etal-2023-meal, agrawal-etal-2023-context}. For example, recent studies have investigated effects of such sample selection strategies for in-context learning~\cite{zhang-etal-2022-active, li-qiu-2023-finding}. However, \textit{for augmentation scenarios, an investigation of sample selection strategies effects is completely lacking}.

The goal of this paper is to compare existing sample selection strategies in few-shot text augmentation. This is measured by the performance of classification models trained on the augmented data. We investigate the typical \textit{paraphrasing} scenario, but also less covered \textit{generation of new samples}. Along with more frequent in-distribution evaluation, we also evaluate on out-of-distribution data. We run our experiments for various LLMs and classification tasks. We identify the best performing sample selection strategy in each scenario (parameter combination) and compare it against two baselines: (1) the zero-shot augmentation, and (2) the few-shot augmentation with random sample selection. We formulate the following research questions:

\begin{description}[labelwidth = 24pt, leftmargin = !]
    \item[RQ1:] \emph{Considering downstream classifier performance, which sample selection strategy performs the best most consistently? (when considering both in-distribution and out-of-distribution setups).}
    \item[RQ2:] \emph{Considering downstream classifier performance, when and how often do the best performing sample selection strategies outperform the baseline strategies?}
\end{description}

We compared 8 different sample selection strategies against 2 baselines strategies (zero-shot and randomly selected samples) on 3 different LLMs (Llama-3.1, Mistral-v0.3 and Gemma-2). We experimented with 7 different datasets (with tasks of sentiment analysis, news classification, question topic classification and natural language inference) with both in-distribution and out-of-distribution splits on RoBERTa as our classifier. Furthermore, we also investigated the \textit{composition of the examples} from the point of labels (whether it is more beneficial to include samples only from the target label being augmented or also from other labels). We investigated two augmentation techniques: \textit{paraphrasing} of existing samples and \textit{generation} of completely new samples. We repeated the whole process 3 times with different random seeds, ensuring the robustness of our results.

The most prominent findings are: 
1) None of the existing sample selection strategies is consistently better than baseline in majority of cases for in-distribution,
2) Selecting examples at random yields the best performance in majority of cases and does not require additional overhead,
3) For out-of-distribution, the \textit{synthetic samples dissimilarity} selection strategy yields the highest performance more often than the baseline strategies, and can be considered for uses where overhead selection costs are not an issue.



\section{Related Work: LLM-based Text Augmentation}

Soon after their advent, new LLMs such as GPT-4 or Llama, started to be used as data augmentation tools, leveraging their ability to produce diversity of texts. The LLM-based augmentation is typically done through paraphrasing~\cite{cegin2024effectsdiversityincentivessample, dai2023auggpt, sen-etal-2023-people}. Less often, LLMs are used to create semantically new samples adhering to a given label~\cite{ubani2023zeroshotdataaug}. LLM-based augmentation has been used for a variety of augmentation tasks such as automated scoring~\cite{fang2023using}, low-resource language generation~\cite{ghosh-etal-2023-dale}, sentiment analysis ~\cite{piedboeuf-langlais-2023-chatgpt, ubani2023zeroshotdataaug, ONAN2023101611}, hate speech detection~\cite{sen-etal-2023-people}, news classification~\cite{piedboeuf-langlais-2023-chatgpt}, content recommendation~\cite{contect-based-recom}, and health symptoms classifications~\cite{dai2023auggpt}.

Recent studies have also used few-shot learning as part of the augmentation by supplying the LLM with various examples from the dataset in the prompts. It has been leveraged for named entity recognition~\cite{ye2024llmdadataaugmentationlarge}, classification performance~\cite{cegin2024effectsdiversityincentivessample} or text summarization~\cite{sahu2024mixsummtopicbaseddataaugmentation}. While the performance of the few-shot approaches in augmentation seems to outperform zero-shot ones (where no examples are used)~\cite{piedboeuf2024evaluationprotocolsdataaugmentation}, the effects of various sample selection strategies are under-explored, as many studies simply select the samples randomly. Only few studies explored other strategies, such as the recent~\cite{cegin2024effectsdiversityincentivessample}, which used a human-inspired sample selection strategy.

While sample selection strategies have found their usage in various in-context learning tasks (significantly altering the performance of LLMs) and while some studies already hint at increased performance of few-shot augmentation over zero-shot augmentation~\cite{cegin2024effectsdiversityincentivessample, piedboeuf2024evaluationprotocolsdataaugmentation}, an investigation of various sample selection strategies for LLM-based augmentation methods is completely lacking.


\section{Study Design}
\label{sec:data_coll}

\begin{figure*}[t]
    \centering
    \includegraphics[width=16cm]{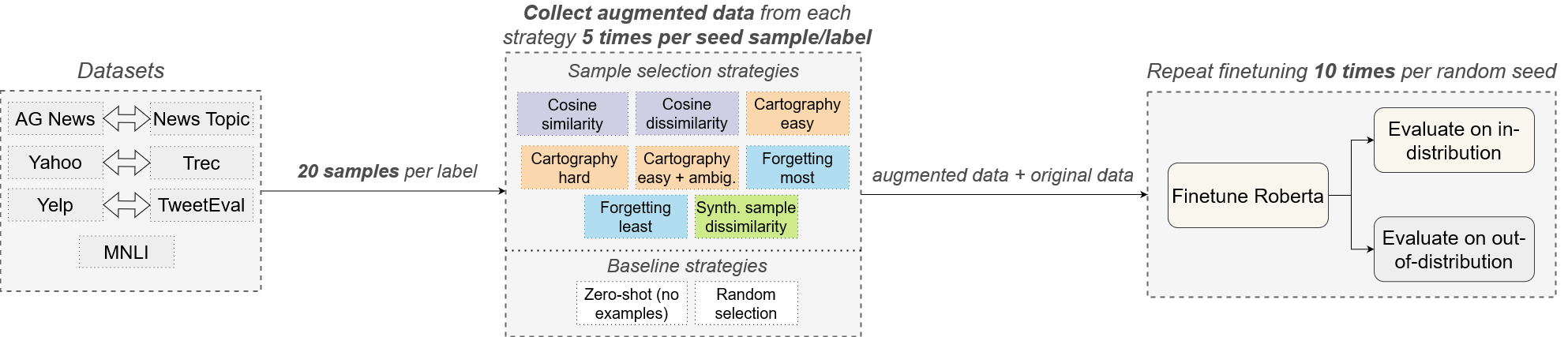}
    \caption{Overview of our methodology. For each dataset, we randomly sample 20 samples per label which are then used to collect up to 5 augmented samples per each seed sample. These seeds are used for fine-tuning with the augmented samples to evaluate each sample selection strategy. This entire process is repeated 3 times with different random seeds. Similar sample selection strategies have the same colour.}
    \label{fig:dataset_build}
\end{figure*}

To assess which sample selection strategies work best for LLM-based data augmentation, we performed a comparative study. The same basic scenario was used in each case: given a text classification dataset, 20 seed samples were selected from each class. For each seed sample, a given LLM ``augmented'' the samples 5 times. This was repeat for each sample selection strategy and type of augmentation technique used (paraphrasing or creating completely new samples). 
Next, a downstream classifier was fine-tuned on both sub-sampled data and augmented samples, and then evaluated on in-distribution and out-of-distribution data. For in-distribution data, we used the original test splits of each dataset, while for the out-of-distribution data we used test splits from a different dataset with the same task (e.g. \textit{Yelp} dataset test split was used as out-of-distribution data when evaluating performance on the \textit{Tweet Eval} dataset for sentiment analysis). 

This scenario was repeated for all sample selection strategies and baselines for a variety of parameters (see below). Then, the performance of the classifiers (measured by F1-macro) was compared for each sample selection strategy to answer the RQ1. This was followed by comparing the best performing sample selection strategies against the best performing baseline strategy of either zero-shot (no examples provided) or randomly selected examples to answer RQ2. We publish all of our results, the code and data used~\footnote{Data and code at \url{https://github.com/kinit-sk/selec-strats-for-aug}}. 

We used a broad range of study parameters to ensure robustness of our results by using both the baseline strategies and the sample selection strategies in a variety of cases. We include the different augmentation techniques and example compositions in terms of labels to capture a wide variety of cases. The whole process was repeated 3 times with different seed samples selected. The study had following parameters:
\begin{itemize}
    \setlength\itemsep{0em}
    \item 8 sample selection strategies (Forgetting with 2 variations, Cartography with 3 variations, Cosine similarity/dissimilarity and Synthetic samples dissimilarity) with 2 2 baseline strategies (zero-shot with no examples provided and random few-shot with examples selected randomly),
    \item 3 LLMs used as augmenters (LLama-3.1-8B, Gemma-2-9B and Mistral-v0.3-7B),
    \item 7 datasets used (\textit{MNLI, Yelp, Tweet sentiment evaluation, AG News, News Topic, Yahoo, Trec}),
    \item 2 types of \textit{composition of examples} used (examples used only from the target label or examples selected from all labels in the dataset),
    \item 2 augmentation techniques (either paraphrasing of existing samples or generation of new label adhering samples),
\end{itemize}

\subsection{Sample Selection Strategies}
\label{sec:established}

We used on the best-performing sample selection strategies identified by previous studies on sample selection in in-context learning~\citep{pecher2024automatic, li-qiu-2023-finding, chang-jia-2023-data, toneva2019empiricalstudyexampleforgetting, zhang-plank-2021-cartography-active}. We used 5-shots per label for each of the sample selection strategies. 

First, we used the \textit{Similarity} and \textit{Dissimilarity} selection that are currently the most popular selection strategies for in-context learning~\citep{an-etal-2023-skill, liu-etal-2022-makes, chang-jia-2023-data}. To select the samples, we calculated the cosine similarity between the feature representation of the samples and then select either the most similar or the most dissimilar ones. In case of paraphrasing, we calculated the similarity towards the sample we were augmenting. In case of generation, we first randomly select one sample and then calculate the similarity towards this sample.

Second, we used the \textit{Synthetic samples dissimilarity} sample selection~\citep{cegin2024effectsdiversityincentivessample}. To select the samples, we first use the LLM to generate a set of synthetic samples and then use the \textit{dissimilarity} selection to select the set of examples from this set.

Third, we used the \textit{Forgetting} strategy that selects the samples based on how often they are forgotten~\citep{toneva2019empiricalstudyexampleforgetting}. To select the samples, we first trained the classifier on the underlying task for a fraction of the overall epochs and observed the training dynamics. For each sample, we calculated how often it is incorrectly classified after already being correctly classified in the previous epoch. Afterwards, these forgetting events are used to select the samples. We explored two different setting in our experiments and choose samples accordingly: 1) \textit{Forgetting most}, where we selected the samples that were the most often forgotten; and 2) \textit{Forgetting least}, where we selected the samples that were forgotten the least amount of times.

Finally, we used the \textit{Cartography} sample selection that measures how easy or hard it is to learn the different samples~\citep{swayamdipta-etal-2020-dataset, zhang-plank-2021-cartography-active}. This ease of learning is determined by training the classifier on the underlying task for a fraction of the overall epochs and looking at the average confidence/probability of the correct class and the variance of this confidence. The samples with high confidence and low variance are considered to be the \textit{easy} to learn samples. At the same time the samples with small confidence and small to medium variance are considered the \textit{hard} to learn ones. The remaining samples are considered to be \textit{ambiguous} (medium confidence or samples with high variance). We explored three different settings in our experiments and choose the samples accordingly: 1) \textbf{Easy} samples, where we sorted the samples based on confidence and choose the top 5 samples with highest confidence for each class; 2) \textbf{Hard} samples, where we sorted the samples based on confidence and choose the bottom 5 samples for each class (i.e., the lowest confidence samples); 3) \textbf{Easy + Ambiguous}, where we first calculated average confidence, selected the samples whose confidence is higher than the average, and then randomly sampled from them.



\subsection{Datasets}
\label{sec:datasets}

For a diverse evaluation, we selected 7 datasets representing tasks of sentiment analysis, news classification, question topic classification and natural language inference. We used  the \emph{News Category}~\cite{misra2022news, misra2021sculpting} and \emph{AG news}~\cite{zhang2015character} for news classification, \emph{Yahoo}~\cite{NIPS2015_250cf8b5} and \emph{Trec}~\cite{li-roth-2002-learning} for question topic classification, MNLI dataset for natural language inference and \emph{TweetEval}~\cite{rosenthal2017semeval} and \emph{Yelp}~\cite{zhang2015character} for sentiment classification. All datasets were multi-class and English. For in-distribution evaluation of classifiers, we used the test split of each dataset. For out-of-distribution evaluation for each dataset, we used the test split of the dataset that is within the same domain, e.g. we used the test split from \textit{Yelp} for \textit{TweetEval} and vice versa (with exception of MNLI, which has its own out-of-distribution test split). While still of the same task, we considered these splits out-of-distribution due to them being collected from other domains or sources (e.g. sentiment analysis of Yelp reviews for classifier trained on tweets). To achieve this evaluation we preprocessed the data by selecting only some of the labels from each dataset and aggregated some labels together. We only generated or paraphrased hypotheses for MNLI given the premise from the dataset. Details about labels used and preprocessing can be found in Appendix~\ref{sec:appendix_dataset_details}. 

\subsection{Evaluation Process}

We randomly selected 20 samples per label from each dataset and repeated this three times with different random seeds. We chose 20 samples per label as this number of seed samples per label should yield the highest effect for augmentation~\cite{cegin2024llmsvsestablishedtext}.  We then augmented the entire selected subset of the dataset for each combination of augmentation technique (\textit{paraphrasing} or \textit{generation}), sample selection strategy (including baselines), augmenting LLM, and \textit{composition of the examples} from the point of labels. We instructed the LLM to collect 5 new samples per one seed sample for each combination of parameters. Prompt templates, specific versions of LLMs used and parameters used for the LLMs can be found in Appendix~\ref{sec:appendix_params_and_prompts}. We did not check the validity of the collected samples, as previous works have already shown that the validity of LLM augmentation methods is quite high~\cite{cegin-etal-2023-chatgpt, cegin2024effectsdiversityincentivessample}.    

We used RoBERTa-base for fine-tuning and used the version of the model from Huggingface. The best working hyperparameters were found via hyperparameter search and these can be found in Appendix~\ref{sec:appendix_finetuning_details}. We trained each model 10 times per each random seed and augmentation parameter combination. The models were trained separately on the data collected from Llama-3.1, Gemma-2 and Mistral. Finally, we computed the F1-macro of all fine-tuned classifiers to allow the comparison of sample selection strategies between themselves and against the baseline strategies.

\section{Study Results}

Our study has multiple parameter dimensions which together yielded more than 1,300 combinations. To keep the results presentation manageable, we aggregated the results for each of the used LLMs and do not distinguish between results from each of them. During our analysis we did not identify any LLM biased towards one of the sample selection strategies.

To keep the comparison of various sample selection strategies simple, we only compare the best performing sample selection strategy combination on the dataset given the augmentation techniques of either \textit{generation} or \textit{paraphrasing} and \textit{composition of labels} in terms of labels. We use the same setting also for the baseline strategies of zero-shot and random few-shot. We wish to identify strategies that provide the best performance most consistently (in the most cases) and outperform the baselines the most.  We analyse the different the augmentation techniques and \textit{composition of labels} and how they influence the classifier performance in Appendix~\ref{sec:appendix_exampl_comp_aug_tech}.

We distinguish between the best performing sample selection strategy for in-distribution data and out-of-distribution data for each of the datasets. To identify the best performing  sample selection strategy (including the baselines) in these cases we compute the mean of the classifier performance across all of the random seeds and compare these means. There were a total of 63 cases for 7 datasets, 3 different LLMs and different 3 random seeds used. After identifying the best performing sample selection strategy, we statistically tested its distribution of classifier performance against the best performing baseline strategy (either zero-shot or random few-shot based on their mean) using Mann-Whitney-U tests with \emph{p=0.05} to measure the number of times the sample selection strategies are statistically significantly better than the best baseline strategy.

\begin{table*}[t!]
    \centering
    \small
    \setlength\tabcolsep{5pt}
    \begin{tabular}{@{}lc|c||c|c||c|c||c|c||c|c||c|c||c|c||c|c@{}}
    \toprule
    \textsc{Dataset}$\rightarrow$ & \multicolumn{2}{c}{\textsc{AG News}} & \multicolumn{2}{c}{\textsc{NewsTopic}} & \multicolumn{2}{c}{\textsc{Yahoo}}  & \multicolumn{2}{c}{\textsc{Trec}} & \multicolumn{2}{c}{\textsc{TweetEval}} & \multicolumn{2}{c}{\textsc{Yelp}} & \multicolumn{2}{c}{\textsc{MNLI}} & \multicolumn{2}{c}{\textsc{\underline{\textbf{TOTAL}}}}\\
    Strategy$\downarrow$  & ID & OD & ID & OD & ID & OD & ID & OD & ID & OD & ID & OD & ID & OD & ID & OD\\
     \midrule
    Zero-shot  & 0 &  0 & 0 &  0 & 0 &  1 & 0 &  \textbf{3} & 0 &  \textbf{6} & 0 &  0 & 0 &  0 & 0 & 10 \\
    Random & \textbf{5} &  2 & \textbf{4} &  1 & 2 &  0 & 2 &  0 & \textbf{2} &  1 & \textbf{3} &  0 & 2 &  2 & \textbf{20} & 6 \\
    \midrule
    Cos. sim. & 1 &  0 & 2 &  \textbf{4} & \textbf{5} &  \textbf{3} & \textbf{5} &  0 & 0 &  0 & \textbf{3} &  0 & 0 & 0 & 16 & 7 \\
    Cos. dissim. & 0 &  0 & 0 &  0 & 2 &  1 & 0 &  2 & 0 &  0 & 0 &  0 & 2 &  2 & 4 & 5 \\
    Forgetting most & 2 &  0 & 0 &  1 & 0 &  \textbf{3} & 0 &  0 & 1 &  1 & 0 &  1 & 0 &  0 & 3 & 6 \\
    Forgetting least & 0 &  1 & 2 &  1 & 0 &  0 & 1 &  0 & \textbf{2} &  0 & 0 &  \textbf{3} & 0 &  0 & 5 & 5 \\
    Carto. hard & 0 &  1 & 0 &  1 & 0 &  1 & 0 &  0 & 1 &  0 & 2 &  \textbf{3} & 0 &  1 & 3 & 7\\
    Carto. easy+amb. & 0 &  0 & 0 &  1 & 0 &  0 & 0 &  0 & \textbf{2} &  0 & 1 &  0 & 2 &  0 & 5 & 1\\
    Carto. easy & 0 &  1 & 0 &  0 & 0 &  0 & 0 &  1 & 1 &  0 & 0 &  0 & 0 &  0 & 1 & 2\\
    Synth. dissim. & 1 &  \textbf{4} & 1 &  0 & 0 &  0 & 1 &  \textbf{3} & 0 &  1 & 0 &  2 & \textbf{3} &  \textbf{4} & 6 & \textbf{14} \\
     \bottomrule
    \end{tabular}
    \caption{No. cases for each sample selection strategy including baseline strategies where each strategy performed the best for each dataset for in-distribution (ID) and out-of-distribution (OD) data. The last \textit{Total} column aggregated all cases for that specific strategy. In total, only the \textit{Synthetic samples dissimilarity} strategy on out-of-distribution  outperforms the baseline strategies most often, while the random few-shot baseline strategy works best for in-distribution.}
    \label{tab:best_perf_strategies}
\end{table*}

\begin{figure*}[!t]
\begin{tabular}{cc}
  \includegraphics[width=0.425\textwidth]{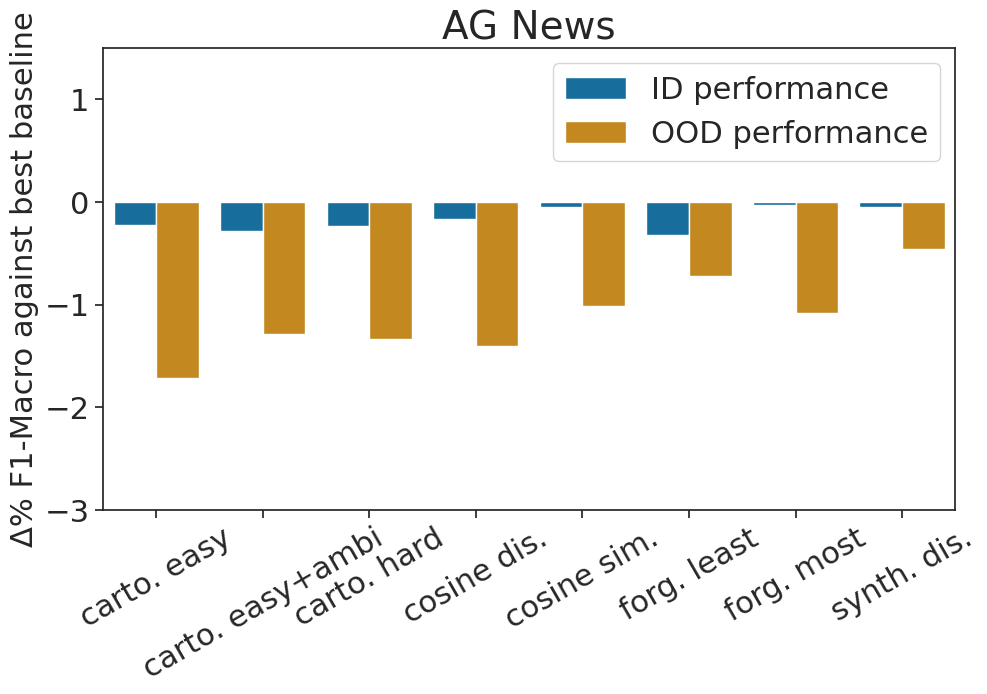} &   \includegraphics[width=0.425\textwidth]{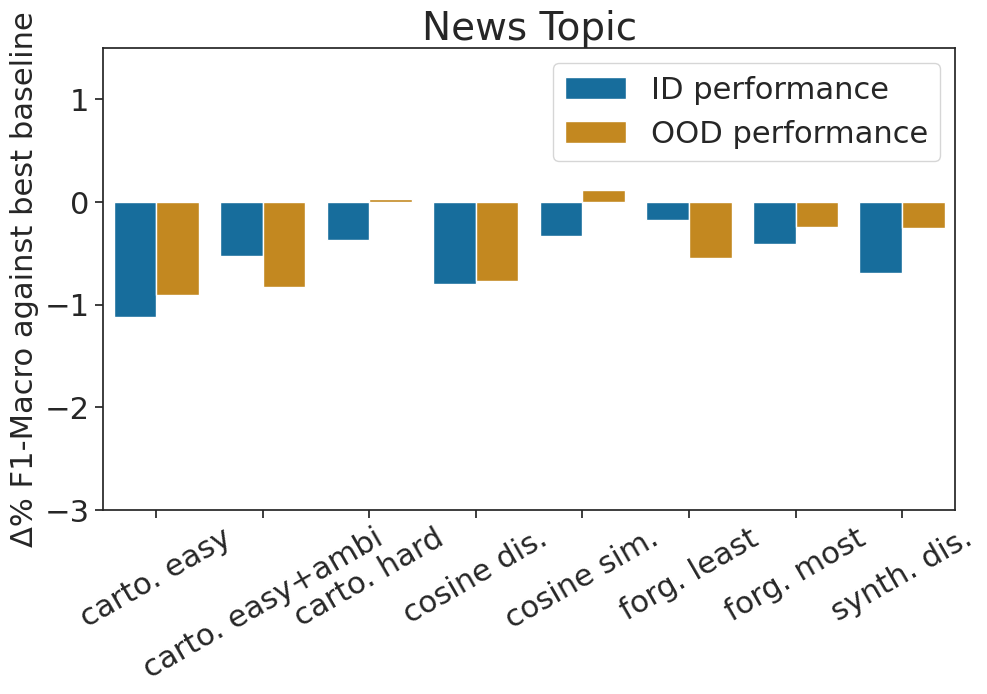}  \\ \includegraphics[width=0.425\textwidth]{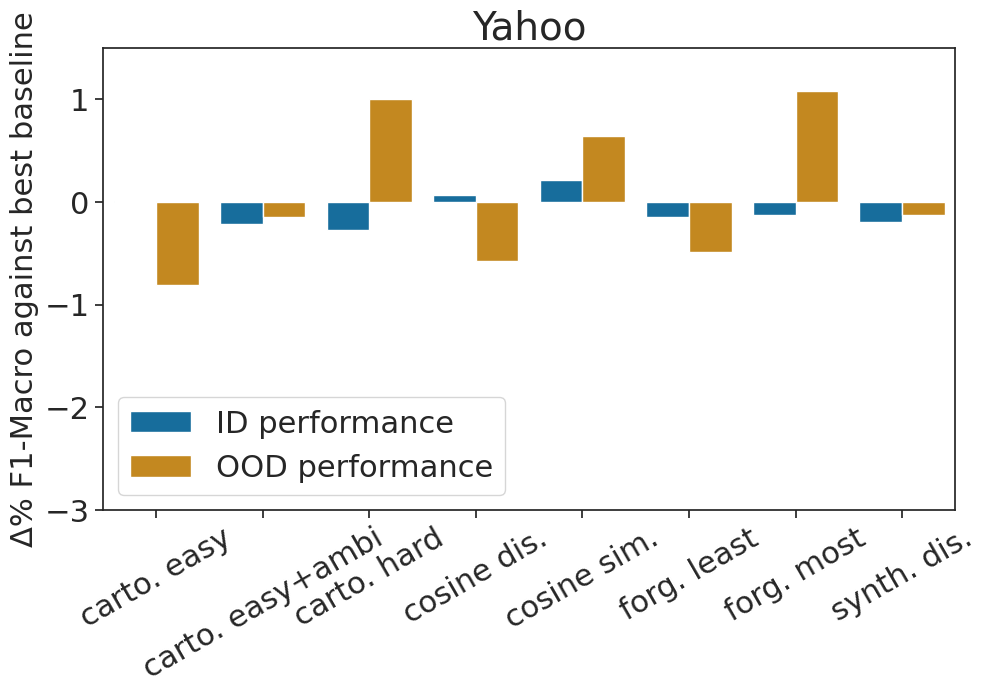} &
  \includegraphics[width=0.425\textwidth]{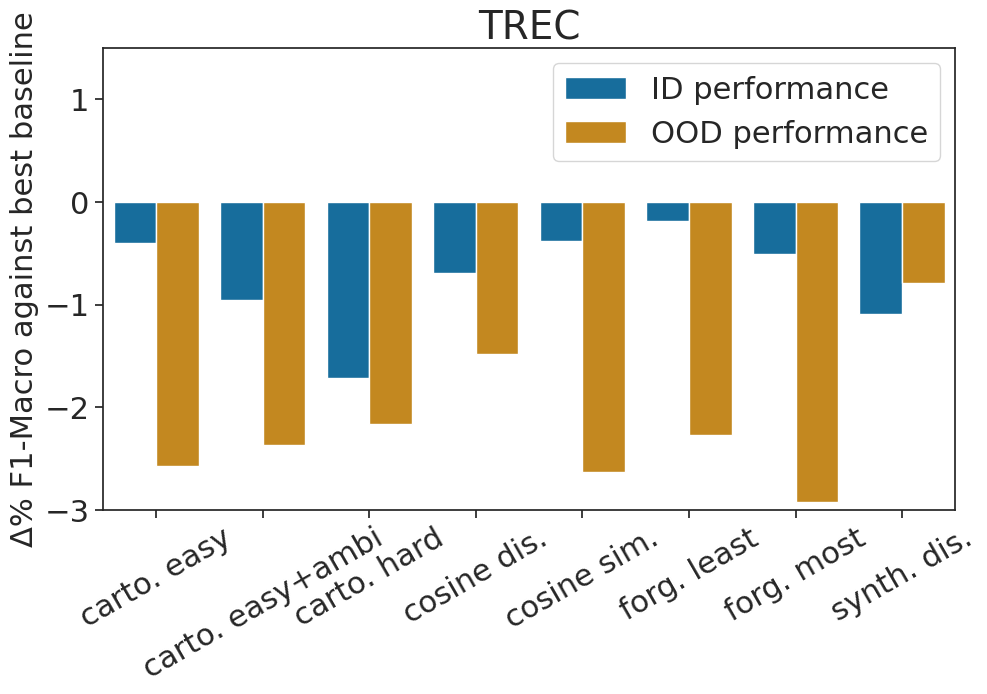} \\ \includegraphics[width=0.425\textwidth]{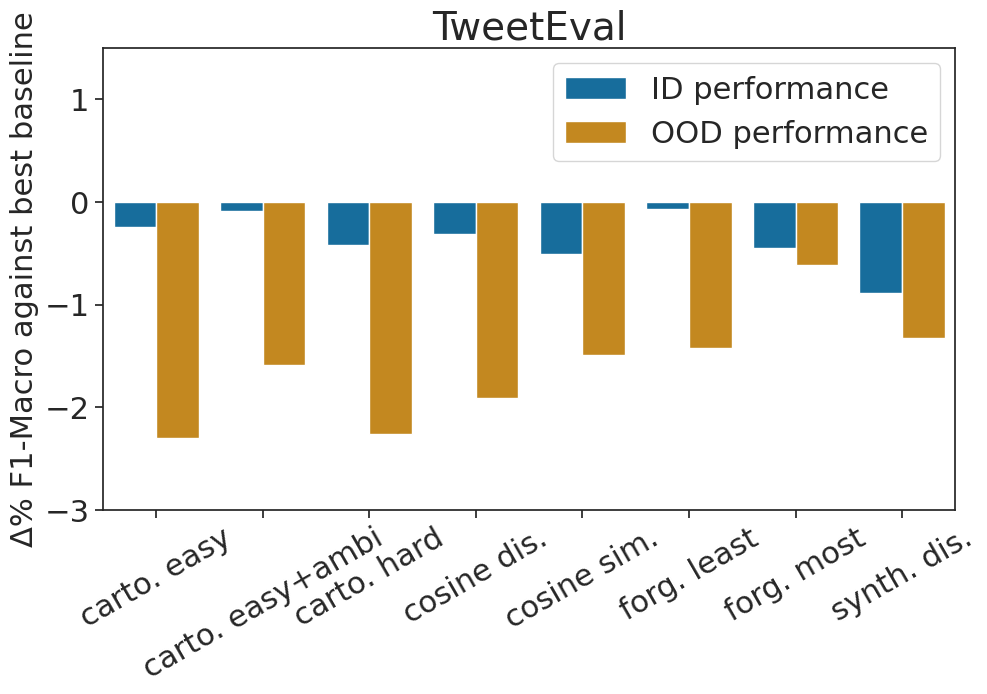} &   \includegraphics[width=0.425\textwidth]{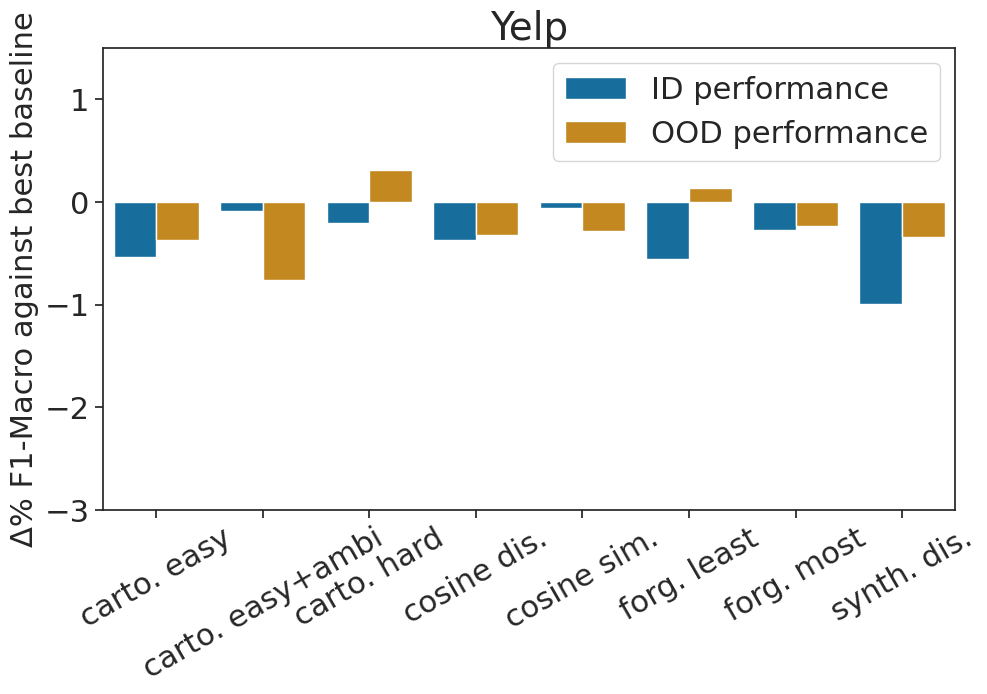} \\
\multicolumn{2}{c}{\includegraphics[width=0.425\textwidth]{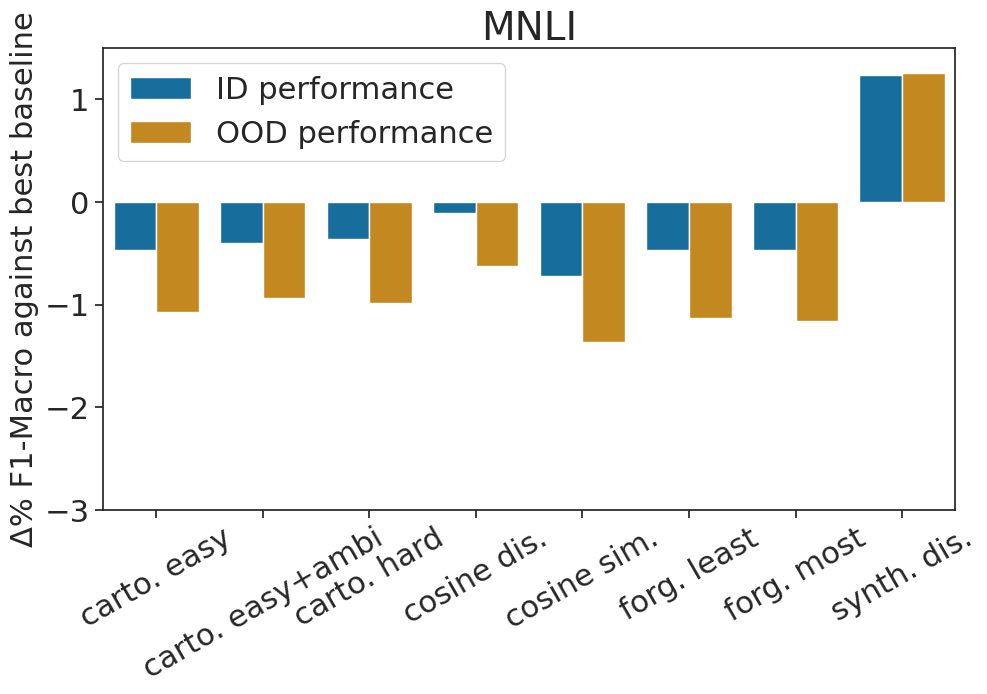}}
\end{tabular}
\caption{Aggregated difference across all LLMs and random seeds in mean F1-Macro for classifiers trained on various sample selection strategies against the best performing baseline of either random few-shot or zero-shot. While some strategies perform well in certain cases as per Table~\ref{tab:best_perf_strategies}, they fail to make a positive impact on classifier performance against baseline strategies in general.}
\label{fig:mean_diff_strategies_vs_baselines}
\end{figure*}

\subsection{Best Performing Sample Selection Strategies}\label{sec:established_newer_strict_comp}

The number of times where each sample selection strategy (including baseline strategies) performed the best for each dataset for in-distribution (ID) and out-of-distribution (OD) data can be found in Table~\ref{tab:best_perf_strategies}. The comparison excluding baseline strategies can be found in Appendix~\ref{sec:appendix_add_results}, together with the performance distributions for each sample selection strategy and dataset. There is no apparent strategy that performed the best across all datasets for both in-distribution and out-of-distribution classifier performance. However, certain sample selection strategies did perform best overall for given data distributions - the \textit{Cosine similarity} strategy performed the best most often from all sample selection strategies (excluding baseline strategies) for in-distribution data in 16 out of 63 cases (25.4\%) and the \textit{Synthetic samples dissimilarity} strategy performed the best most often for out-of-distribution data in 14 out of 63 cases (22.22\%).

Some of the strategies seem to be biased for certain datasets, performing well in those cases. For example, the \textit{Synthetic samples dissimilarity} strategy is well suited for the \textit{MNLI} dataset for both in-distribution and out-of-distribution cases, while the \textit{Cosine similarity} strategy works best for \textit{Yahoo} dataset for both data and on \textit{Trec} dataset for in-distribution data.

Considering the sample selection strategies without the baselines, the \textit{Cosine similarity} strategy performs best for in-distribution data in 18 out of 63 cases (28.57\%), followed by the \textit{Synthetic samples dissimilarity} strategy in 12 out of 63 cases (19.05\%). For out-of-distribution comparison of strategies, the best strategy is the \textit{Synthetic samples dissimilarity} strategy in 22 out of 63 cases (34.92\%) followed by the \textit{Cartography with hard samples} strategy in 9 out of 63 cases (14.29\%).

We answer the \emph{RQ1} as follows: Considering the sample selection strategies without the baseline strategies, the most effective sample selection strategy is \textit{Cosine similarity} for in-distribution data and \textit{Synthetic samples dissimilarity} for out-of-distribution data. However, we also note that in certain cases both of these strategies fail to perform as the best strategy even once (\textit{TweetEval} for \textit{Cosine similarity} and \textit{Yahoo} for \textit{Synthetic samples dissimilarity}).

\subsection{Comparison of Best Sample Selection Strategies Against Baseline Strategies}\label{sec:comparison_best_samples_vs_baselines}

We compare the best identified sample selection strategies from Section~\ref{sec:established_newer_strict_comp} against baseline strategies as per Table~\ref{tab:best_perf_strategies} and also provide aggregated difference across all cases in mean F1-Macro for various sample selection strategies against the best performing baseline of either random few-shot or zero-shot in Figure~\ref{fig:mean_diff_strategies_vs_baselines}.

For in-distribution classifier performance we identified as the best performing sample selection strategy the \textit{Cosine similarity} performing best in 16 out of 63 cases (25.4\%). The best performing baseline on in-distribution data is random few-shot which achieved best performance in 20 out of 63 cases (31.75\%), an increase by 4 cases compared to the \textit{Cosine similarity} strategy. Out of the 16 cases where the \textit{Cosine similarity} performed best, it was statistically significantly better than the best baseline strategy in 9 cases. The random few-shot baseline also achieved best performance in a variety of cases across all the datasets, which the \textit{Cosine similarity} strategy did not and was outperformed by the \textit{Cosine similarity} strategy only on the \textit{Yahoo} and \textit{Trec} datasets.

For out-distribution classifier performance we identified as the best performing sample selection strategy the \textit{Synthetic samples dissimilarity} performing best in 14 out of 63 cases (22.22\%). The best performing baseline on in-distribution data is zero-shot which achieved best performance in 10 out of 63 cases (15.87\%), performing worse than the \textit{Synthetic samples dissimilarity} strategy in 4 cases. Out of the 14 cases where the \textit{Synthetic samples dissimilarity} performed best, it was statistically significantly better than the best baseline strategy in 6 cases. While the \textit{Synthetic samples dissimilarity} works well only on some datasets (achieving no best cases for \textit{NewsTopic} and \textit{Yahoo} datasets), the same can be said about both the baselines: zero-shot strategy achieves no best cases on 4 datasets and random few-shot achieves no best cases on 3 datasets.

All of the sample selection strategies fail to make a consistent impact on model performance as can be seen in Figure~\ref{fig:mean_diff_strategies_vs_baselines}. While there are cases where increases are apparent in both in-distribution and out-of-distribution performance (on the \textit{MNLI} dataset), the sample selection strategies fail to outperform consistently the best baseline of either zero-shot with no examples or randomly selected samples for few-shot, as the increase in performance on one random seed with specific seed samples is mitigated by losses of performance on another random seed with different seed samples.  

We answer the \emph{RQ2} as follows: When comparing in-distribution performance, the best baseline of random few-shot strategy performs better than the best sample selection strategy of \textit{Cosine similarity} in 4 more cases. Additionally, the random few-shot strategy works well across all dataset for in-distribution classifier performance. When comparing out-of-distribution classifier performance, the best sample selection strategy of \textit{Synthetic samples dissimilarity} performs better than the best baseline of zero-shot in 4 more cases. Neither the baselines nor the \textit{Synthetic samples dissimilarity} strategy perform well on all datasets.

\section{Discussion}
The results of our experiments lead to the following observations: First, the \textit{Cosine similarity} strategy was best among sample selection strategies for in-distribution classifier performance. The inclusion of most similar samples in the prompts as examples during augmentation may bias the LLM for the distribution of the dataset even more, thus resulting in this increased performance.

Second, the \textit{Synthetic samples dissimilarity} strategy was best among sample selection strategies for out-of-distribution classifier performance. The strategy used in~\cite{cegin2024effectsdiversityincentivessample} was inspired by crowdsourcing methods~\cite{larson-etal-2020-iterative} that focused specifically on collecting data for better out-of-distribution performance. This indicates that the design of this method, where previously collected samples are used as examples based on their dissimilarity, translates well for LLM-based augmentation.

Third, when comparing the best sample selection strategies against baseline strategies, the \textit{Cosine similarity} strategy is outperformed by the random few-shot selection strategy. Not only does the random few-shot strategy perform better more often, but it does so across multiple datasets, while the \textit{Cosine similarity} strategy fails to perform best even once for some datasets. This hinders the applicability of this method, where it is clearly outperformed in some cases by baseline strategies or other sample selection strategies, as it does not always guarantee good performance for in-distribution data.

Fourth, the \textit{Synthetic samples dissimilarity} strategy outperforms the baseline strategies for out-distribution performance, but it does not so across all datasets. However, neither the random few-selection strategy or the zero-shot approach perform well on all datasets. This implicates that increasing performance across all out-of-distribution cases for all datasets is difficult problem which neither of the sample selection strategies can achieve.

Fifth, as seen in Figure~\ref{fig:mean_diff_strategies_vs_baselines} the aggregated increase of classifier performance when using sample selection strategies is small or negative, indicating that sample selection strategies do not work well for all random seeds. Given the increased costs of using sample selection strategies this result favours the baseline strategies for text augmentation in general.

Sixth, comparing the baseline strategies between themselves, the random few-shot selection performs the best on in-distribution classifier performance, while the zero-shot strategy only ever performs well for out-of-distribution classifier performance. This might be due to the LLMs getting biased towards the examples provided and thus being more likely to produce augmentations that follow the distribution of the seed samples more closely, resulting in increased in-distribution classifier performance. However, this might not be robust enough for good out-of-distribution classifier performance.

To summarise, while the \textit{Synthetic samples dissimilarity} strategy outperforms the baseline strategies for out-of-distribution classifier performance, the baseline strategies outperform the sample selection strategies for in-distribution classifier performance. However, any increase for both in-distribution or out-of-distribution classifier performance is marginal and brings with itself increased costs for collecting augmentation by using sample selection strategies. While there are cases where sample selection strategies work best, they do not so consistently. This underlines the need for better sample selection strategies for LLM-based text augmentation.
 
\section{Conclusion}

We compared the effects of prominent sample selection strategies of few-shot learning for LLM-based text augmentation scenario. We evaluated the downstream classifier performance on in-distribution and out-of-distribution data. We compared selection strategies against 2 baseline strategies (random few-shot and zero-shot). This comparison was done using 3 different LLMs, 7 different datasets, and 2 augmentation techniques (paraphrasing and new sample generating). 

Our comparison indicates that the baseline strategies outperform sample selection strategies for in-distribution performance. For out-of-distribution performance the \textit{Synthetic sample dissimilarity} strategy is best in more cases than the baseline strategies. However, the improvements are marginal and are not present in all datasets. Given the increased computations needed to use these sample selection strategies and their lacklustre performance, the baseline strategies represent a good default option for few-shot augmentation practitioners. The space for better sample selection strategies for LLM-based text augmentation remains open.

\section*{Limitations}
    We note several limitations to our work. 
    
    First, we only used datasets, augmentation methods, and LLMs for the English language and did not investigate cases of multi-lingual text augmentation.

    Second, we did not use various patterns of prompts  and followed those used in previous studies~\cite{cegin-etal-2023-chatgpt, larson-etal-2020-iterative}. Different prompts could have effects on the quality of text augmentations, but they would also radically increase the size of this study, and thus, we decided to leave this for future work and focused on simplest prompts possible.

    Third, we did not use newer LLMs for classification fine-tuning via PEFT methods (e.g fine-tuning of Llama-3 or Mistral using QLoRA). While such inclusion would strengthen our findings, we decided not to use these models as evaluation of these models is very costly and takes a long time due to their size, which results in them being mostly used with a small subset of the testing data~\cite{chang-jia-2023-data, li-qiu-2023-finding, gao-etal-2021-making, koksal-etal-2023-meal}. This, in return, can lead to unintentionally cherry picked results. We see the usage of such fine-tunings as the extension of our work left for future work.

    Fourth, for the LLM augmentation methods we used only Llama-3.1-8B, Mistral-v0.3-7B and Gemma-2-9B. We did not use larger models (e.g. 70B versions) as their increased performance in text augmentation for model accuracy has been shown~\cite{cegin2024effectsdiversityincentivessample} to be not that significant when compared to variants of LLMs with fewer parameters, while the inference costs compared to these smaller models is much higher.

    Fifth, we used only 5-shots on 20 seeds per label selected on each dataset. While a bigger number of seeds and shots could have been used, we opted for smaller numbers to keep the study manageable and the cost of the study low. In addition, previous study~\cite{pecher2024automatic} found that sample selection is more impactful when choosing only a small set of samples and using more samples does not necessarily lead to better results due to the limited context size of the models. Furthermore, obtaining larger annotated datasets (e.g., hundreds of samples per class) is not feasible for many domains in practice. As such, our findings are beneficial even for these domains. The exploration of additional number of shots and seeds is an interesting direction that can be explored in the future.

    Sixth, we only focus on classification tasks and make no claims about the effects of sample selection strategies used in LLM-based text augmentation on other NLP tasks.

    Seventh, we do not know if any of the 6 datasets used in this study have been used for training the LLMs we used for data collection and if this had any effect on our results and findings. As such, we do not know how much would be the comparison of established and newer LLM augmentation methods different on new, unpublished datasets. This limitation is part of the recently recognised possible ``LLM validation crisis'', as described by~\cite{li2023task}.

    Eight, we used only one feature representation model for the sample selection strategies that required similarity or dissimilarity of samples and the usage of different feature representation models could alter the performance of these sample selection strategies.

\section*{Acknowledgments}

This work was partially supported by Modermed, a project funded by the Slovak Research and Development Agency, GA No. APVV-22-0414. The work was also partially funded by the EU NextGenerationEU through the Recovery and Resilience Plan for Slovakia under the projects No. 09I03-03-V03-00020 and No. 09I01-03-V04-00006.

This work was supported by the Ministry of Education, Youth and Sports of the Czech Republic through the e-INFRA CZ (ID:90254).
\bibliography{compressed}

\appendix

\section{Ethical considerations}
Based on a thorough ethical assessment, performed on the basis of intra-institutional ethical guidelines and checklists tailored to the use of data and algorithms, we see no ethical concerns pertaining directly to the conduct of this research. Albeit production of new data through LLMs bears several risks, such as introduction of biases, the small size of the produced dataset, sufficient for experimentation is, at the same time, insufficient for any major machine learning endeavours, where such biases could be transferred.

We follow the license terms for all the models and datasets we used (such as the one required for the use of the Llama-3.1 model) – all models and datasets allow their use as part of research.

\section{Classifier fine-tuning details}\label{sec:appendix_finetuning_details}

We selected the best hyper-parameters after using hyper-parameter search. We used the same batch size across all datasets using 64 batch size, used \emph{2e-5} learning rate, dropout 0.2, maximum amount of tokens (512) trimmed and padded, and 50 number of epochs. We used AdamW optimizer in all cases.

\section{Dataset details}\label{sec:appendix_dataset_details}

As we did not use all of the dataset labels and samples in each of the dataset, we list our setup here. All used datasets are in English language. We either aggregated or relabelled the classes we used in datasets to ensure that datasets from all domains of sentiment analysis, news classification and question topic classification had the same classes. This made the out-of-distribution evaluation much easier. 

We used all the labels for the \textit{TweetEval} dataset and for the \textit{Yelp} dataset we aggregated and relabelled the \textit{one star} and \textit{two star} labels as \textit{negative}, the \textit{three star} as \textit{neutral} and the \textit{four star} and \textit{five star} labels as \textit{positive}. 

We used all the labels of the \textit{AG News} dataset and for the \textit{News Topic} dataset we aggregated and relabelled the \textit{WORLD NEWS, POLITICS} as \textit{U.S. NEWS}  as \textit{World}, \textit{SCIENCE, TECH} as Science and Technology and additionally also used samples with labels \textit{Sports} and \textit{Business}. 

For the \textit{Yahoo} dataset we used labels \textit{Society \& Culture, Science \& Mathematics, Health, Education \& Reference, Sports, Business \& Finance}. We used only some labels of the Trec dataset and mapped them to the \textit{Yahoo} dataset labels in the following way by aggregation and relabelling: on the \textit{Society \& Culture} label we mapped the \textit{HUM:gr, HUM:ind, NUM:date, HUM:desc, ENTY:religion} labels, on the \textit{Science \& Mathematics} label we mapped  the \textit{ENTY:animal,  NUM:volsize, ENTY:plant, NUM:temp} labels, on the \textit{Health} label we mapped  the \textit{ENTY:body, ENTY:dismed} labels, on the \textit{Education \& Reference} label we mapped  the \textit{ABBR:abb, DESC:def, DESC:desc} labels, on the \textit{Sports} label we mapped  the \textit{ENTY:sport} label and on the \textit{Business \& Finance} label we mapped  the \textit{ENTY:cremat} label.

Finally, we used all the labels in the \textit{MNLI} dataset.

\section{Prompts and parameters used for LLM-based augmentation}\label{sec:appendix_params_and_prompts}

For all of the LLMs used during augmentation we used the same parameters: maximum amount of new tokens set to 1024, sampling enabled, with \textit{top p} set to 1 and \textit{temperature} set to 1. We used 4-bit quantisation for faster and cheaper inference on all LLMs and used intrustion-tuned versions for each of the LLMs. Specifically, we used Mistral-v0.3-7B-instruct~\footnote{https://huggingface.co/mistralai/Mistral-7B-Instruct-v0.3}, Llama-3.1-8B-Instruct~\footnote{https://huggingface.co/meta-llama/Llama-3.1-8B-Instruct} and Gemma-2-9B-Instruct~\footnote{https://huggingface.co/google/gemma-2-9b-it}. We collected 1 response and asked the LLMs to produce 5 augmentations per seed or label of that seed.

We used different prompts for generating new samples and paraphrasing existing samples. These prompts were also varied based on the dataset used.

For paraphrasing with few-shot we used this prompt: \textit{You will be given examples from '{task}' classification dataset, each labelled with a specific category. Based on the examples, paraphrase a given text 5 times with the '{label}' category. Output each paraphrased text in the form of a numbered list separated by new lines. The text: '{text}'. Examples: {examples}}

For paraphrasing with zero-shot we used this prompt: \textit{You are given a '{task}' classification dataset. Paraphrase a given text 5 times with the '{label}' category. Output each generated text in the form of a numbered list separated by new lines. The text: '{text}'}

For few-shot paraphrasing of the question topic classification datasets we used this prompt: \textit{You will be given examples of questions from '{task}' classification dataset, each labelled with a specific topic. Based on the examples of questions, paraphrase a given question 5 times with the '{label}' topic. Output each paraphrased question in the form of a numbered list separated by new lines. The question: '{text}' Examples: {examples}}

For paraphrasing with zero-shot of the question topic classification datasets we used this prompt: \textit{You are given a '{task}' classification dataset. Paraphrase a given question 5 times with the '{label}' category. Output each generated question in the form of a numbered list separated by new lines. The question: '{text}'}

For few-shot paraphrasing of the MNLI dataset we used this prompt: \textit{You will be given a premise and hypothesis pair together with their label from a Natural Language Inference dataset. Based on the examples, paraphrase 5 times a hypothesis that '{label}' the given premise. The given premise: '{premise}'. Output each paraphrased hypothesis in the form of a numbered list separated by new lines. The hypothesis: '{text}' Examples: {examples}}

For paraphrasing with zero-shot of the MNLI dataset we used this prompt: \textit{You will be given a premise from a Natural Language Inference dataset. Paraphrase 5 times a hypothesis that '{label}' the given premise. The given premise: '{premise}'. Output each paraphrased hypothesis in the form of a numbered list separated by new lines. The hypothesis: '{text}'}

For generating new samples with few-shot we used this prompt: \textit{You will be given examples from '{task}' classification dataset, each labelled with a specific category. Based on the examples, generate 5 new texts that fit the '{label}' category. Output each generated question in the form of a numbered list separated by new lines. Examples: {examples}}

For generating new samples with zero-shot we used this prompt: \textit{You are given a '{task}' classification dataset. Generate 5 new texts that fit the '{label}' category. Output each generated question in the form of a numbered list separated by new lines.}

For few-shot generating new samples of the question topic classification datasets we used this prompt: \textit{You will be given examples of questions from '{task}' classification dataset, each labelled with a specific topic. Based on the examples of questions, generate 5 new questions that fit the '{label}' topic. Output each generated question in the form of a numbered list separated by new lines. Examples: {examples}}

For generating new samples with zero-shot of the question topic classification datasets we used this prompt: \textit{You are given a '{task}' classification dataset. Generate 5 new questions that fit the '{label}' category. Output each generated question in the form of a numbered list separated by new lines.}

For few-shot generating new samples of the MNLI dataset we used this prompt: \textit{You will be given a premise with a label from a Natural Language Inference dataset. Based on the examples, generate 5 new hypotheses that '{label}' the given premise. The given premise: '{premise}'. Output each generated hypothesis in the form of a numbered list separated by new lines. Examples: {examples}}

For generating new samples with zero-shot of the MNLI dataset we used this prompt: \textit{You will be given a premise with a label from a Natural Language Inference dataset. Generate 5 new hypotheses that '{label}' the given premise. The given premise: '{premise}'. Output each generated hypothesis in the form of a numbered list separated by new lines.}

\section{Additional Results and Visualisations for Sample Selection Strategies and Their Effect on Model Performance}\label{sec:appendix_add_results}

\begin{figure*}[!t]
\begin{tabular}{cc}
  \includegraphics[width=0.45\textwidth]{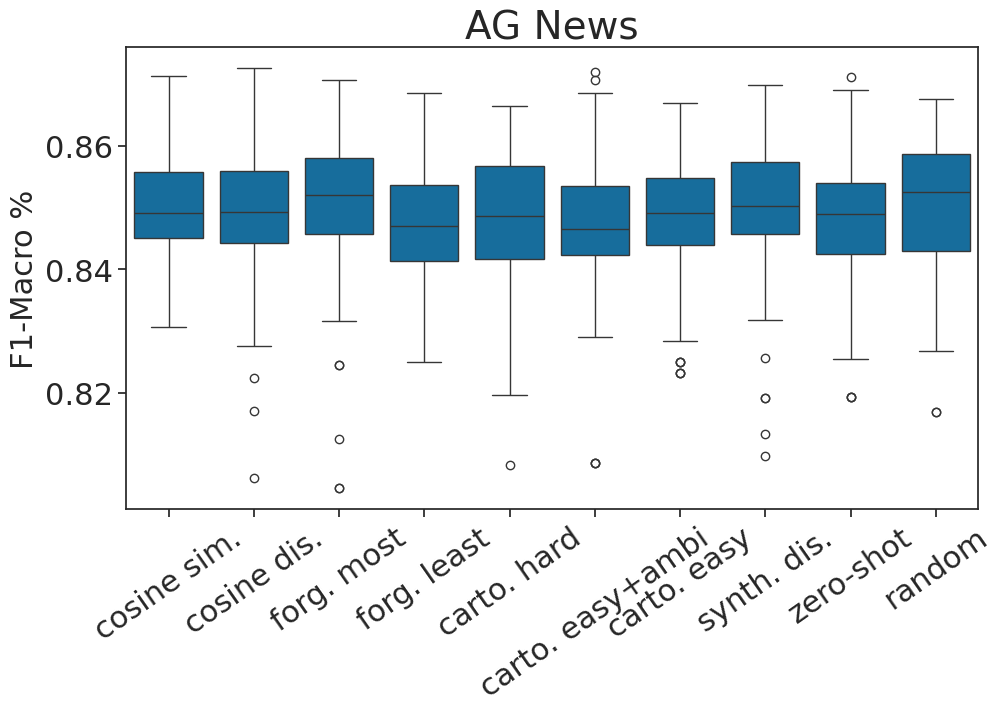} &   \includegraphics[width=0.45\textwidth]{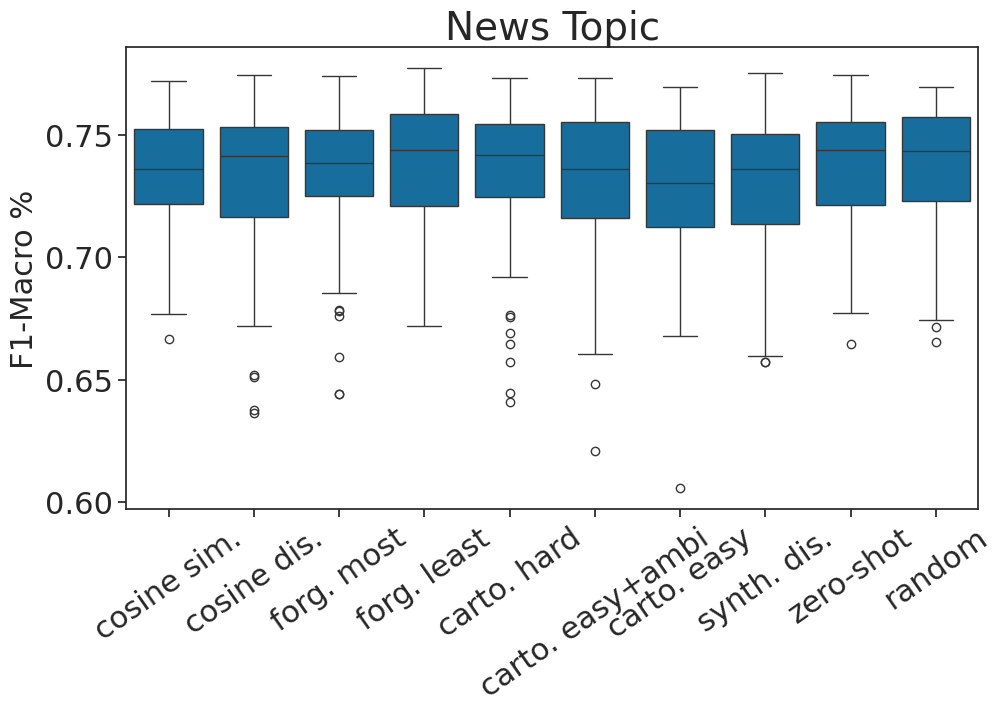}  \\ \includegraphics[width=0.45\textwidth]{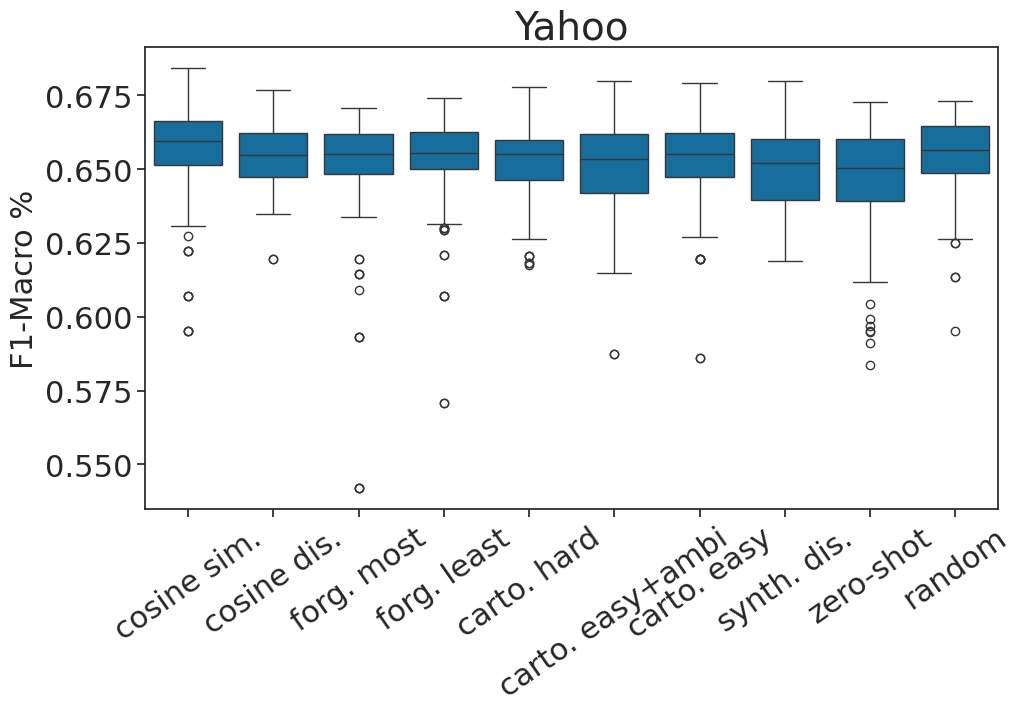} &
  \includegraphics[width=0.45\textwidth]{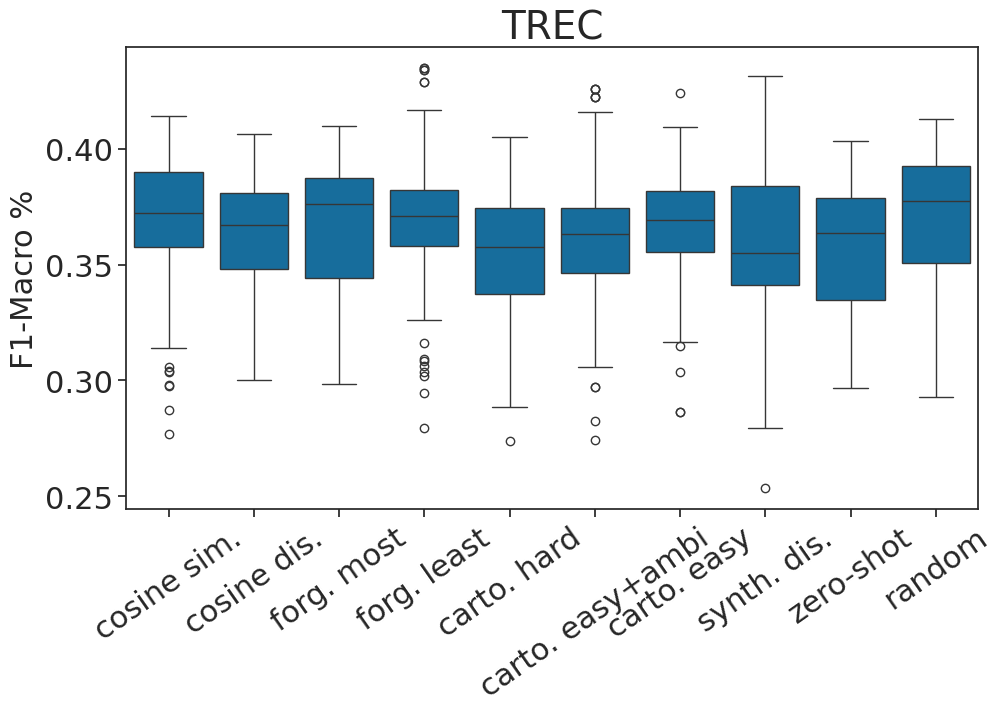} \\ \includegraphics[width=0.45\textwidth]{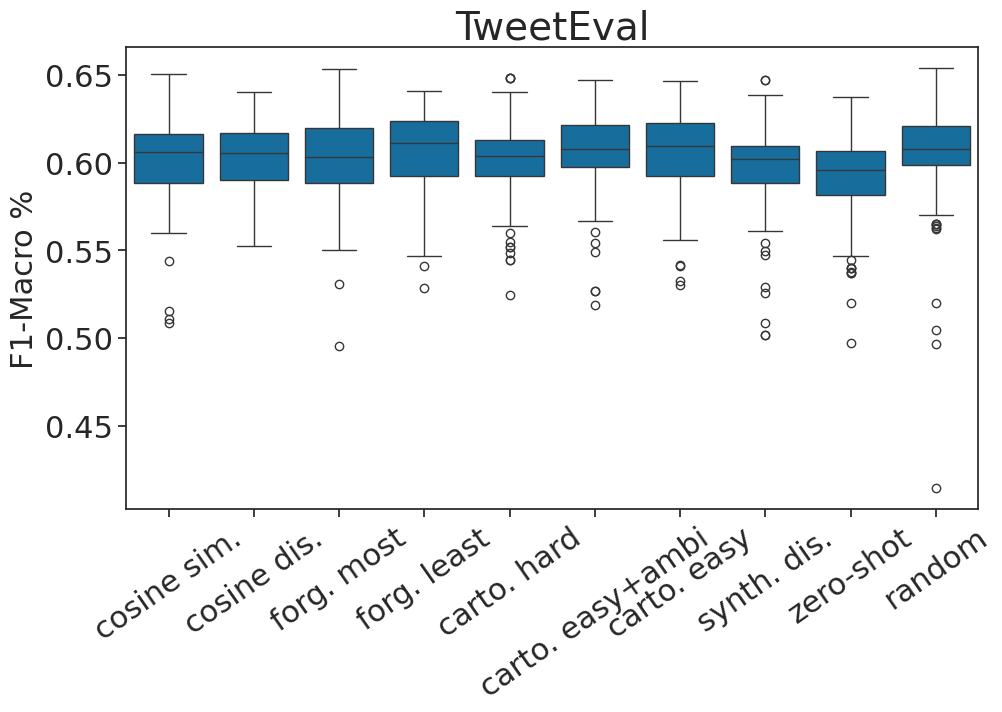} &   \includegraphics[width=0.45\textwidth]{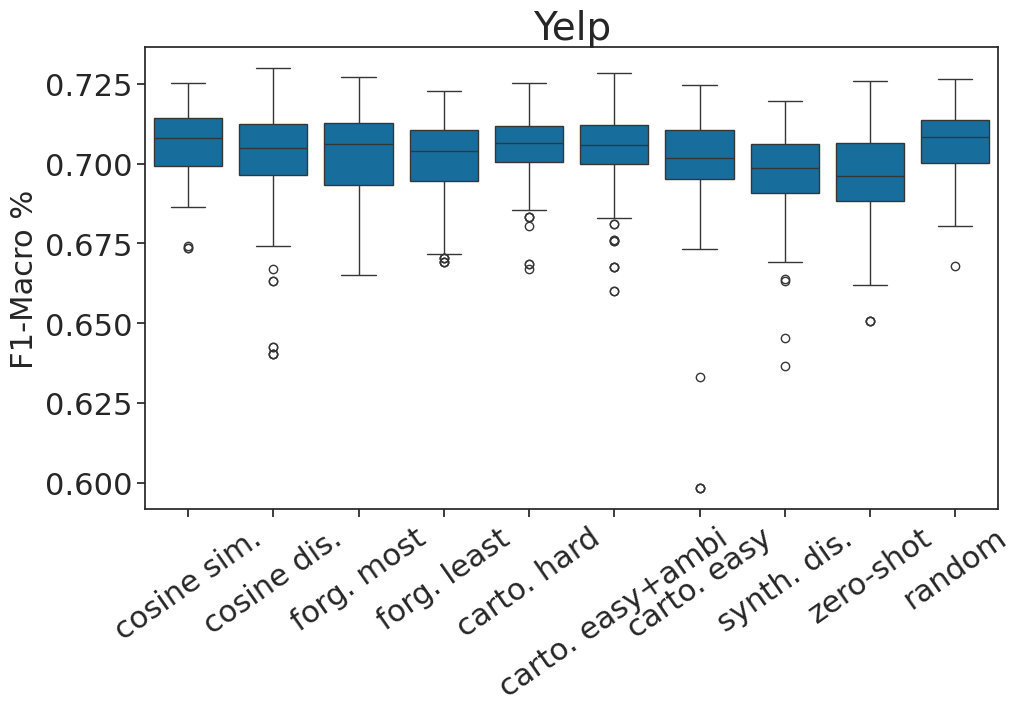} \\
\multicolumn{2}{c}{\includegraphics[width=0.45\textwidth]{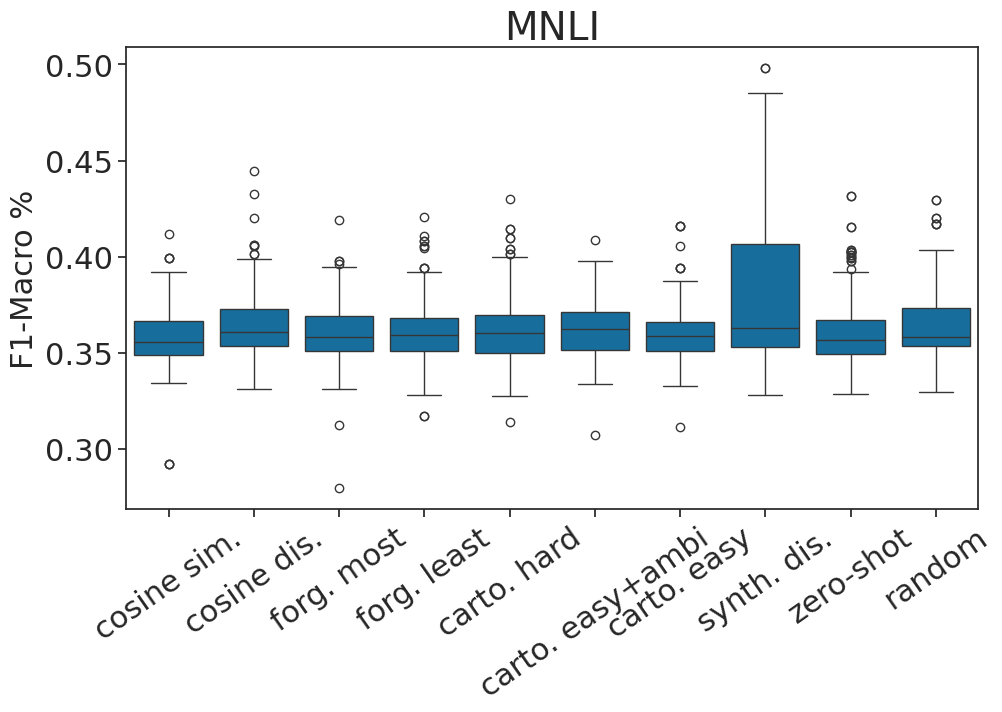}}
\end{tabular}
\caption{Aggregated performance across all LLMs and random seeds in F1-Macro for classifiers trained on various sample selection strategies together with the baselines of either random few-shot or zero-shot on in-distribution data.}
\label{fig:boxplots_perf_ID}
\end{figure*}

\begin{figure*}[!t]
\begin{tabular}{cc}
  \includegraphics[width=0.45\textwidth]{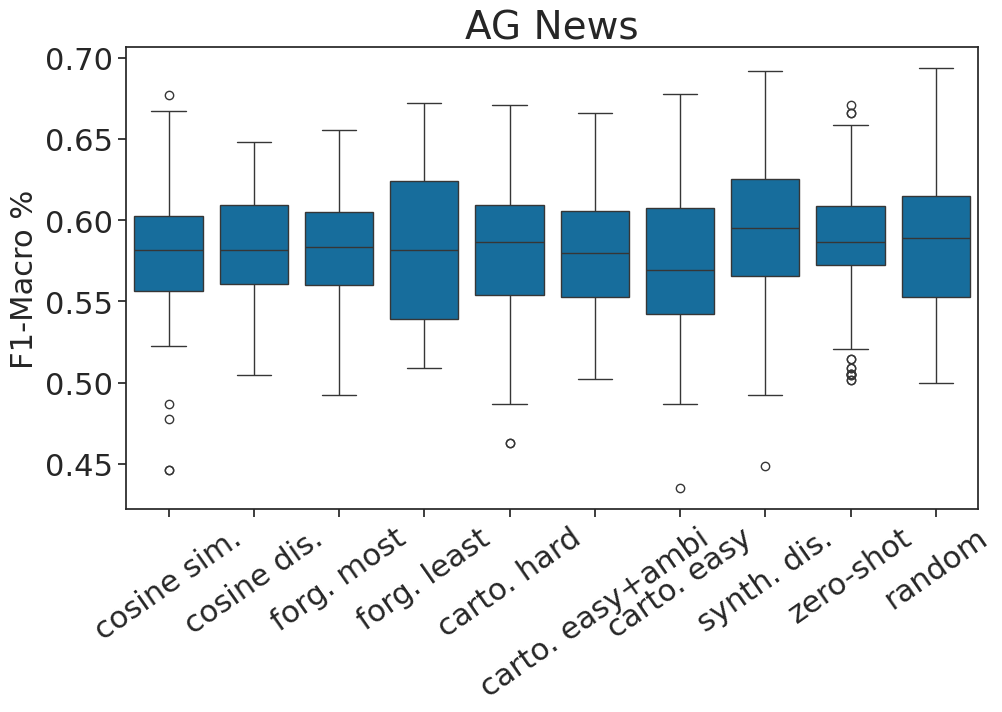} &   \includegraphics[width=0.45\textwidth]{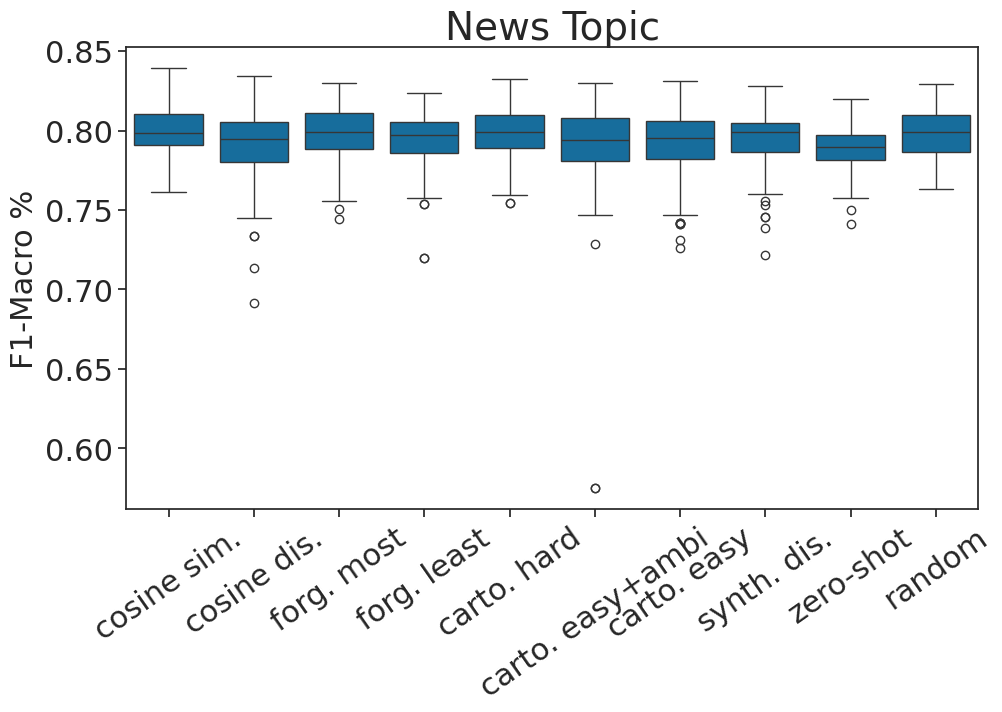}  \\ \includegraphics[width=0.45\textwidth]{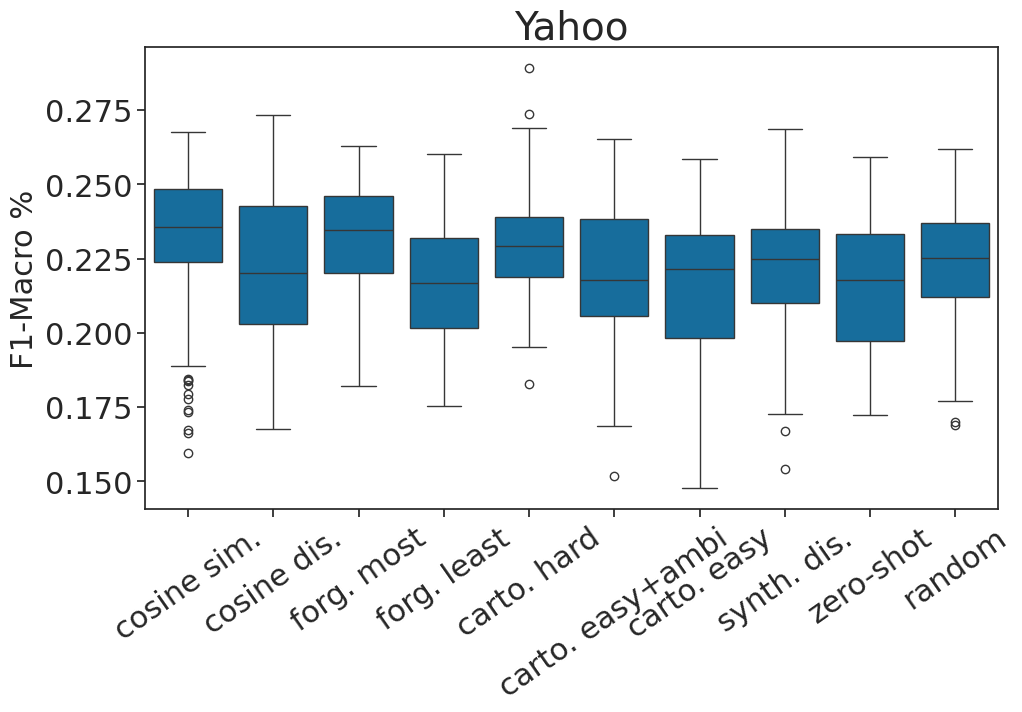} &
  \includegraphics[width=0.45\textwidth]{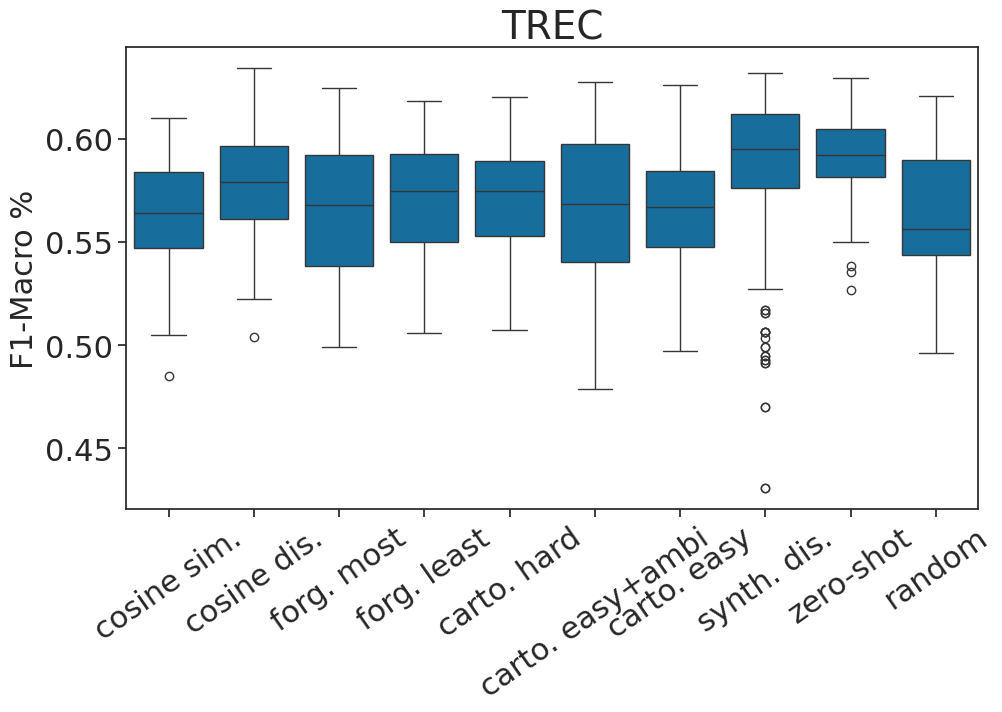} \\ \includegraphics[width=0.45\textwidth]{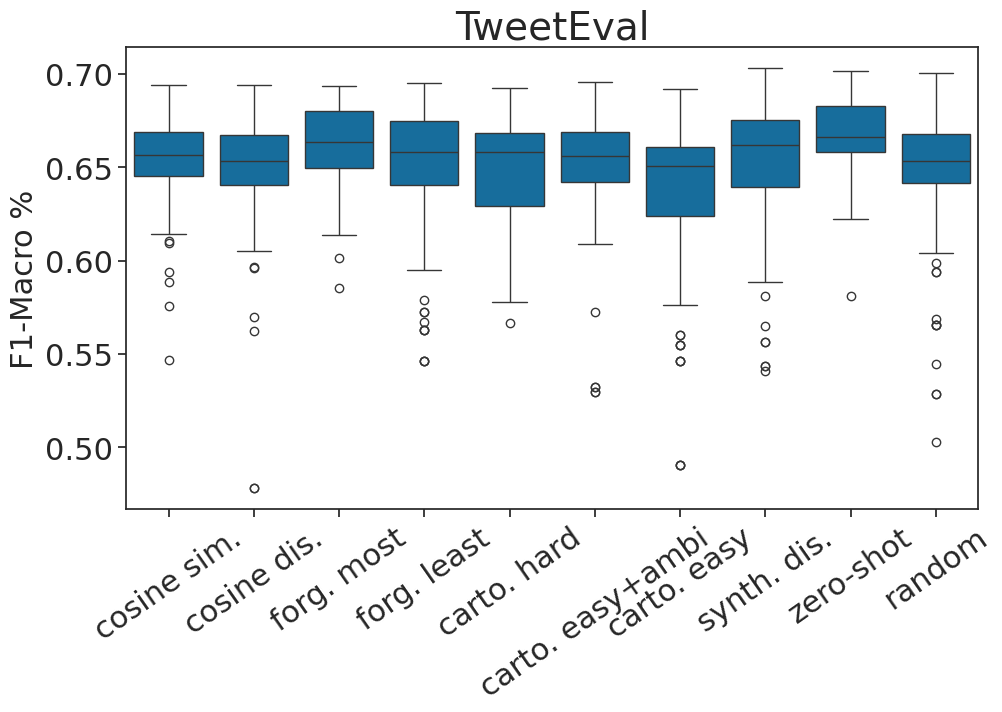} &   \includegraphics[width=0.45\textwidth]{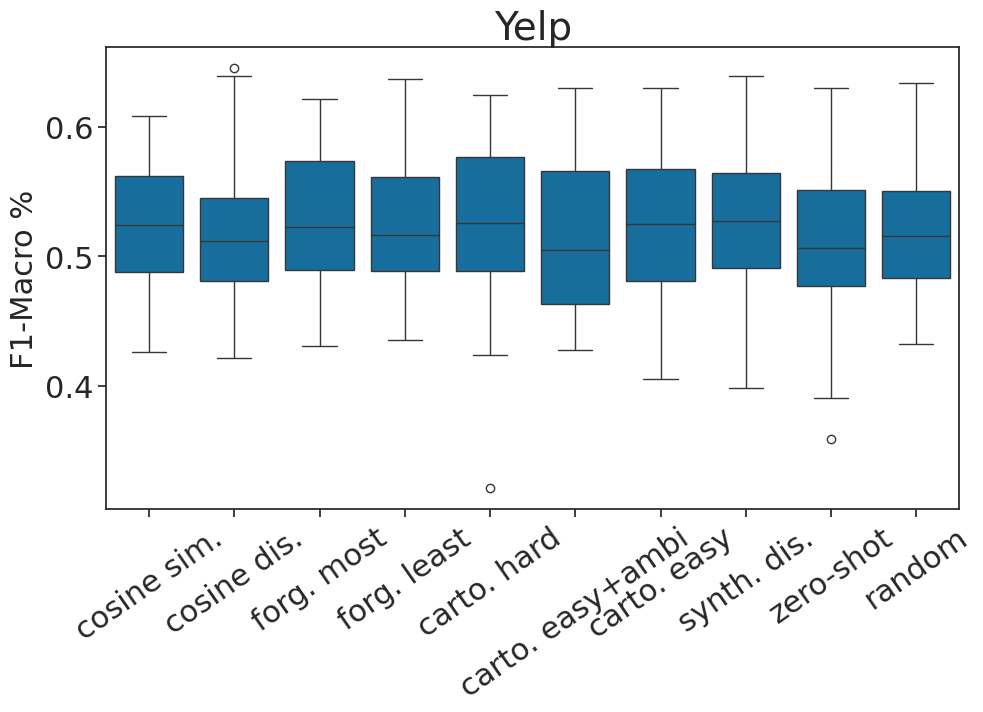} \\
\multicolumn{2}{c}{\includegraphics[width=0.45\textwidth]{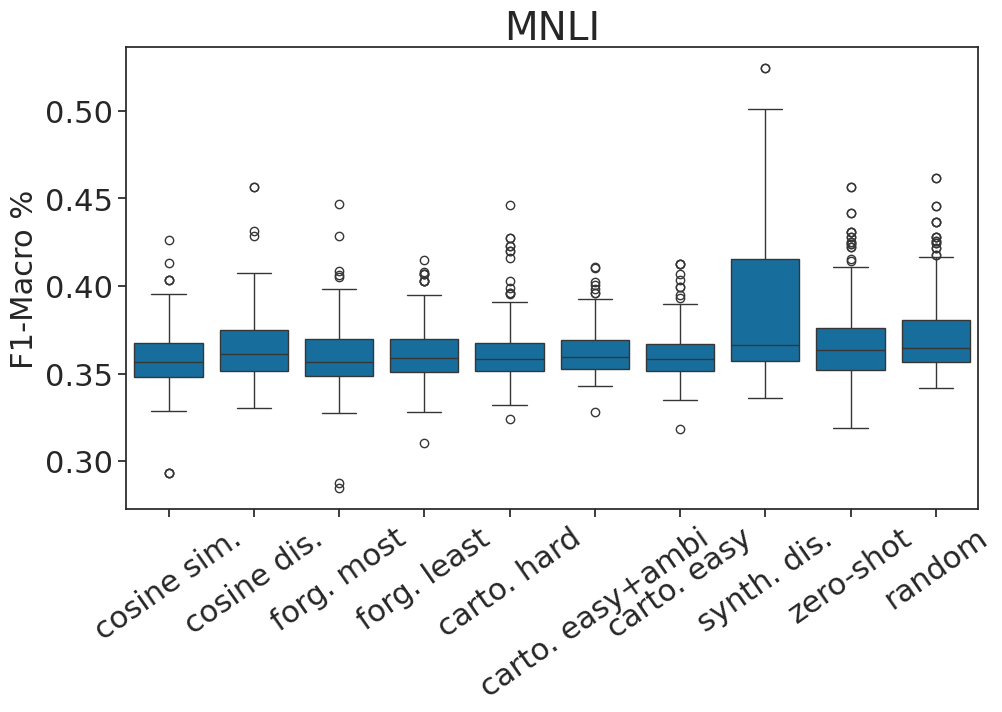}}
\end{tabular}
\caption{Aggregated performance across all LLMs and random seeds in F1-Macro for classifiers trained on various sample selection strategies together with the baselines of either random few-shot or zero-shot on out-of-distribution data.}
\label{fig:boxplots_perf_OOD}
\end{figure*}

We provide the comparison of all sample selection strategies between each other without the baselines in Table~\ref{tab:best_perf_strategies_no_baselines}. Additionally we also provide boxplot visualisation for aggregated performance all LLMs and random seeds in F1-Macro for classifiers trained on various
sample selection strategies together with the baselines of either random few-shot or zero-shot for both in-distribution and out-of-distribution data in Figures~\ref{fig:boxplots_perf_ID} and~\ref{fig:boxplots_perf_OOD}.

\begin{table*}[t!]
    \centering
    \small
    \setlength\tabcolsep{5pt}
    \begin{tabular}{@{}lc|c||c|c||c|c||c|c||c|c||c|c||c|c||c|c@{}}
    \toprule
    \textsc{Dataset}$\rightarrow$ & \multicolumn{2}{c}{\textsc{AG News}} & \multicolumn{2}{c}{\textsc{NewsTopic}} & \multicolumn{2}{c}{\textsc{Yahoo}}  & \multicolumn{2}{c}{\textsc{Trec}} & \multicolumn{2}{c}{\textsc{TweetEval}} & \multicolumn{2}{c}{\textsc{Yelp}} & \multicolumn{2}{c}{\textsc{MNLI}} & \multicolumn{2}{c}{\textsc{\underline{\textbf{TOTAL}}}}\\
    Strategy$\downarrow$  & ID & OD & ID & OD & ID & OD & ID & OD & ID & OD & ID & OD & ID & OD & ID & OD\\
     \midrule
    Cos. sim & 1 &  0 & \textbf{3} &  \textbf{4} & \textbf{6} &  \textbf{3} & \textbf{5} &  0 & 0 &  0 & 3 &  0 & 0 &  0  & \textbf{18} & 7 \\
    Cos. dissim. & 1 &  0 & 1 &  0 & 3 &  1 & 0 &  2 & 0 &  0 & 0 &  0 & 2 &  2 & 7  & 5\\
    Forgetting most & 3 &  0 & 0 &  1 & 0 &  \textbf{3} & 0 &  0 & 1 &  \textbf{4} & 0 &  1 & 0 &  0 & 4 & 9\\
    Forgetting least & 0 &  2 & \textbf{3} &  1 & 0 &  0 & 1 &  0 & \textbf{3} &  1 & 0 &  \textbf{3} & 0 &  0 & 7 & 7 \\
    Carto. hard & 0 &  2 & 0 &  1 & 0 &  1 & 0 &  0 & 1 &  1 & \textbf{4} &  \textbf{3} & 0 &  1 & 5 & 9\\
    Carto. easy+amb. & 0 &  0 & 0 &  2 & 0 &  0 & 0 &  0 & \textbf{3} &  0 & 1 &  0 & \textbf{4} &  0 & 8 & 2 \\
    Carto. easy & 0 &  1 & 0 &  0 & 0 &  0 & 0 &  1 & 1 &  0 & 1 &  0 & 0 &  0 & 2 & 2 \\
    Synth. dissim. & \textbf{4} &  \textbf{4} & 2 &  0 & 0 &  1 & 3 & \textbf{6} & 0 &  3 & 0 &  2 & 3 &  \textbf{6} & 12 & \textbf{22} \\
     \bottomrule
    \end{tabular}
    \caption{No. cases for each sample selection strategy without baseline strategies where each strategy performed the best for each dataset for in-distribution (ID) and out-of-distribution (OD) data. The last \textit{Total} column aggregated all cases for that specific strategy. The \textit{Synthetic samples dissimilarity} strategy perform best on out-of-distribution  classifier performance, while the \textit{Cosine similarity} strategy performs best on in-distribution classifier performance.}
    \label{tab:best_perf_strategies_no_baselines}
\end{table*}

\section{Effects of Composition of Examples and Augmentation Techniques on Classifier Performance}\label{sec:appendix_exampl_comp_aug_tech}

As our study had multiple parameters mentioned in Section~\ref{sec:data_coll}, we additionally also report results for two different parameters used: \textit{composition of examples} based on labels (using only examples from the label under augmentation or using examples from every label in the dataset) and augmentation techniques (using either \textit{paraphrasing} of existing samples or \textit{generation} of new samples). We report results for both parameters in Tables~\ref{tab:augmentation_technique_effect} and~\ref{tab:example_comp_effect}.

\begin{table}[!t]
\centering
\small
\setlength\tabcolsep{3pt}
\begin{tabular}{@{}lcc@{}}
\toprule
\emph{Type of Augmentation} & \emph{Best for ID} & \emph{Best for OD}\\ \midrule
\emph{Generation} & 16 (25.40\%) & \textbf{42 (66.67\%)}  \\
\emph{Paraphrasing} & \textbf{47 (74.60\%)} & 21 (33.33\%)   \\ \bottomrule
\end{tabular}
\caption{No. cases where each type of augmentation performed the best for in-distribution (ID) and out-of-distribution (OD) data. The \textit{generation} augmentation works best for out-of-distribution data, while the \textit{paraphrasing} augmentation works best for in-distribution data.}
\label{tab:augmentation_technique_effect}
\end{table}

\begin{table}[!t]
\centering
\small
\setlength\tabcolsep{3pt}
\begin{tabular}{@{}lcc@{}}
\toprule
\emph{Composition of Examples Type} & \emph{Best for ID} & \emph{Best for OD}\\ \midrule
\emph{Only From Label Under Aug.} & 28 (44.45\%) & 26 (49.06\%)  \\
\emph{From All Labels} & \textbf{35 (55.55\%)} & \textbf{27 (50.94\%)}   \\ \bottomrule
\end{tabular}
\caption{No. cases where each type of \textit{composition of examples} type performed the best for in-distribution (ID) and out-of-distribution (OD) data. While the inclusion of examples from all the labels in the dataset works best, the increase in no. cases is small.}
\label{tab:example_comp_effect}
\end{table}

Each augmentation technique has the best effect on classifier performance for either in-distribution or out-of-distribution as per Table~\ref{tab:augmentation_technique_effect}. For out-of-distribution performance the \textit{generation} of new samples is the best most often, while for in-distribution performance the \textit{paraphrasing} of existing samples works best. Exceptions to this are in the \textit{Yelp} dataset, where \textit{paraphrasing} of existing samples is best for out-of-distribution performance and \textit{generation} of new samples for in-distribution performance.

The difference between \textit{composition of examples} based on labels is much smaller than for augmentation techniques as is shown in Table~\ref{tab:example_comp_effect}. While the inclusion of samples from all labels in the dataset is better more often, the difference is quite small. We noticed that for out-of-distribution performance the inclusion of samples from all labels worked best on question topic classification datasets and \textit{TweetEval} dataset, while the other datasets worked better with only examples from the label under augmentation used. 

\end{document}